\documentclass[10pt,twocolumn,letterpaper]{article}

\usepackage{iccv}
\usepackage{times}
\usepackage{epsfig}
\usepackage{graphicx}
\usepackage{amsmath}
\usepackage{amssymb}
\usepackage[font=small]{subfig}
\usepackage{multirow}
\usepackage{comment}

\usepackage[pagebackref=true,breaklinks=true,letterpaper=true,colorlinks,bookmarks=false]{hyperref}

\iccvfinalcopy % *** Uncomment this line for the final submission

\def\iccvPaperID{3545} % *** Enter the ICCV Paper ID here

\ificcvfinal\fi
\begin{document}

%%%%%%%%% TITLE
\title{Towards Pure End-to-End Learning\\for Recognizing Multiple Text Sequences from an Image}

\author{Zhenlong Xu\textsuperscript{1}\thanks{Zhenlong Xu did most of this work when he was an intern in Hikvision Research Institute.}
	\qquad\qquad
	Shuigeng Zhou\textsuperscript{1}\thanks{Corresponding author.}\qquad\qquad
	Zhanzhan Cheng\textsuperscript{2}\qquad\qquad \\
	Fan Bai \textsuperscript{1}\qquad\qquad
	Yi Niu\textsuperscript{2} \qquad\qquad
	Shiliang Pu\textsuperscript{2}\qquad\qquad\qquad
	\\
	\textsuperscript{1}Shanghai Key Lab of Intelligent Information Processing, and School of \\Computer Science, Fudan University, Shanghai 200433, China\\
	\textsuperscript{2}Hikvision Research Institute, China\\
	{\tt\small \{zlxu18,sgzhou,fbai17\}@fudan.edu.cn}\\
	{\tt\small \{chengzhanzhan,niuyi,pushiliang\}@hikvision.com}
}
%\author{Zhenlong Xu\\
%	Shanghai Key Lab of Intelligent Information Processing, and School of
%	Computer Science, Fudan University, Shanghai 200433\\
%	China\\
%	{\tt\small zlxu18@fudan.edu.cn}
%	\and
%	Second Author\\
%	Institution2\\
%	First line of institution2 address\\
%	{\tt\small secondauthor@i2.org}
%}

\maketitle
%\thispagestyle{empty}

%%%%%%%%% ABSTRACT
\begin{abstract}
	Here we address a challenging problem: recognizing multiple text sequences from an image by pure end-to-end learning. It is twofold: 1) Multiple text sequences recognition. Each image may contain multiple text sequences of different content, location and orientation, and we try to recognize all the text sequences contained in the image. 2) Pure end-to-end (PEE) learning.
	We solve the problem in a pure end-to-end learning way where each training image is labeled by only text transcripts of all contained sequences, without any geometric annotations.
	%Each training image is annotated with only the text transcripts of all sequences contained in the image, but NOT any geometric annotations.
	%location and spatial relationship information about these sequences.
	%We use only the annotated text contents of images to train the recognizer.
	
	Most existing works recognize multiple text sequences from an image in a non-end-to-end (NEE) or quasi-end-to-end (QEE) way, in which each image is trained with both text transcripts and text locations.
	Only recently, a PEE method was proposed to recognize text sequences from an image where the text sequence was split to several lines in the image. However, it cannot be directly applied to recognizing multiple text sequences from an image.
	
	So in this paper, we propose a pure end-to-end learning method to recognize multiple text sequences from an image. Our method directly learns multiple sequences of probability distribution conditioned on each input image, and outputs multiple text transcripts with a well-designed decoding strategy.
	To evaluate the proposed method, we constructed several datasets mainly based on an existing public dataset and
	two real application scenarios. 
	Experimental results show that the proposed method can effectively recognize multiple text sequences from images, and outperforms CTC-based and attention-based baseline methods.
	
	% background
	%In text reading, sequence learning based methods have achieved promising results.
	% previous method
	%These methods transfer a text image into a sequence of characters, which is generally implemented with CTC or attention-based strategies.
	% problems
	%While in real-life scenarios, many images contains more than one text strings, but very limited methods focus on such task. %which relies on the detection stage for generating the cropped text images.
	% details of our method
	%In order to handle such problems, we propose a novel method called {\color{red}XXX} for directly generating multiple sequences of characters from images.
	%{\color{red}XXX} attempts to directly learn the multiple sequences of probability distribution conditioned on the input image, and then outputs multiple text strings with a well-designed decoding strategy.
	%The model with {\color{red}XXX} solves the multiple sequence generation process from text images, and is trained end-to-end with only text-level annotations.
	% experiments
	%According to some demands in real scenarios, we build a set of useful datasets for evaluating the proposed method.
	%Experiments on these datasets and a public dataset {\color{red}XXXX} demonstrate that {\color{red}XXX} can handle the multiple text recognition problem and achieves impressive results.
\end{abstract}

%%%%%%%%% BODY TEXT
\section{Introduction}
% background
Recognizing text from images, evolved from the classical optical character recognition (OCR) problem, has been a popular research topic~\cite{cheng2017focusing,jaderberg2016reading,shi2017end,shi2018aster} in pattern recognition and computer vision areas for more than a decade, due to its wide applications such as handwriting recognition, automatic card reading, and image understanding. Though significant technical progress has been made, this problem is far from being well solved, considering various complex application scenarios.

Existing works~\cite{liao2018textboxes++,liao2017textboxes,wang2011end,wang2010word} of text recognition from images mainly employ a pipeline that contains a text detection module and a text recognition module. The former is used to detect text content (\eg characters, words or text sequences) in images, while the latter is responsible for reading the cropped text images. Technically, we can subsume these works into \emph{non-end-to-end} (NEE for short) methods.
%A common feature of these methods is that they need a \emph{separate} text detector to locate the text content in an image. In essence, what they do is to recognize a single text sequence from an image.
However, given an image containing multiple text sequences, these methods have to
first detect
%{\color{red}first detect}
these sequences, and then recognize them one by one.
%The success of deep leaning in computer vision brings strong power to text recognition from images. A number of deep neural network based methods (including those based on CTC and attention mechanisms) were proposed and achieved a big stride forward in performance.

With the popularity of deep learning, more and more works~\cite{he2018end,li2017towards,liu2018fots} attempt to read text in an ``end-to-end'' way by first pre-training the detection module or recognition module, and then conducting a joint training on the detection and recognition tasks.
%They treat the text recognition problem as a sequence learning problem, where images and texts are separately encoded as patch sequences and character sequences respectively, and train the recognizer in an `end-to-end' way.
%Different from the early NEE methods, they give up the \emph{separate} text detection module, instead, integrate text detection and text recognition into a unified network framework. However, in training the deep networks, they have to annotate the training images with both text \emph{content} and \emph{location} (by bounding box). So we call them \emph{quasi-end-to-end} (QEE) methods.
Different from the early NEE methods, they integrate the detection module and recognition module into a unified network, but still train the whole model with both text transcripts and geometric annotations (\eg bounding boxes). Therefore, we call them \emph{quasi-end-to-end} (QEE) methods.
% li2017towards,liu2018fots

Recently, Bluche \etal~\cite{bluche2017scan} proposed an attention based method to recognize text lines from images in a \emph{pure end-to-end} (PEE for short) way.
We say it is a PEE method in the sense that the training images are annotated with \emph{only} text content, no location information is needed. However, this method can recognize only one text sequence from an image.
Though the target text sequence may split into several lines in the image, they treat the text lines as a whole, and the order of text lines constituting the sequence must be pre-defined.
Essentially, this method can handle only \emph{single sequence recognition} (SSR) problem.

\begin{figure}[t]
	\begin{center}
		\subfloat[]{
			\includegraphics[width=0.4\linewidth]{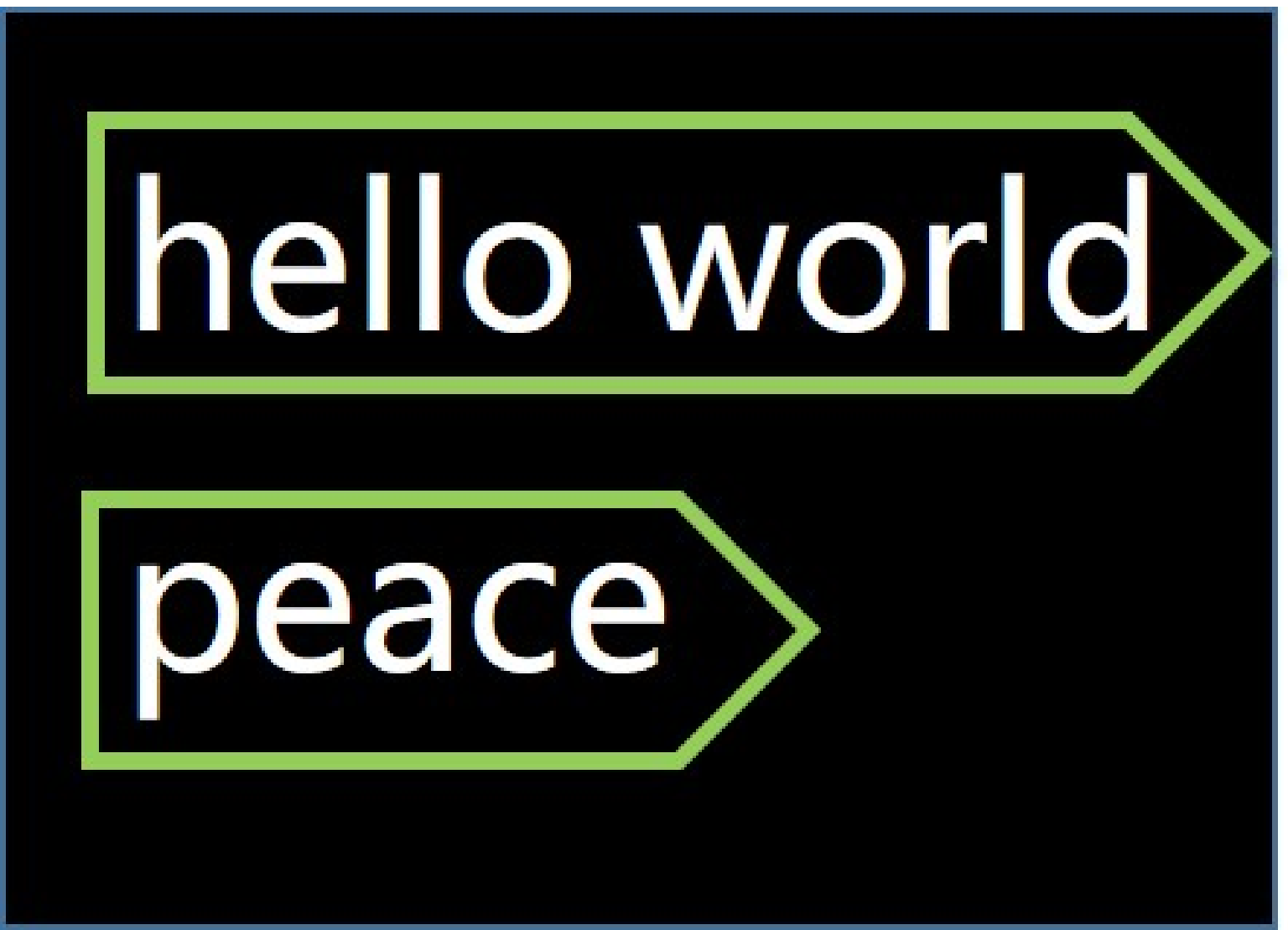}
		}
		\subfloat[]{
			\includegraphics[width=0.56\linewidth]{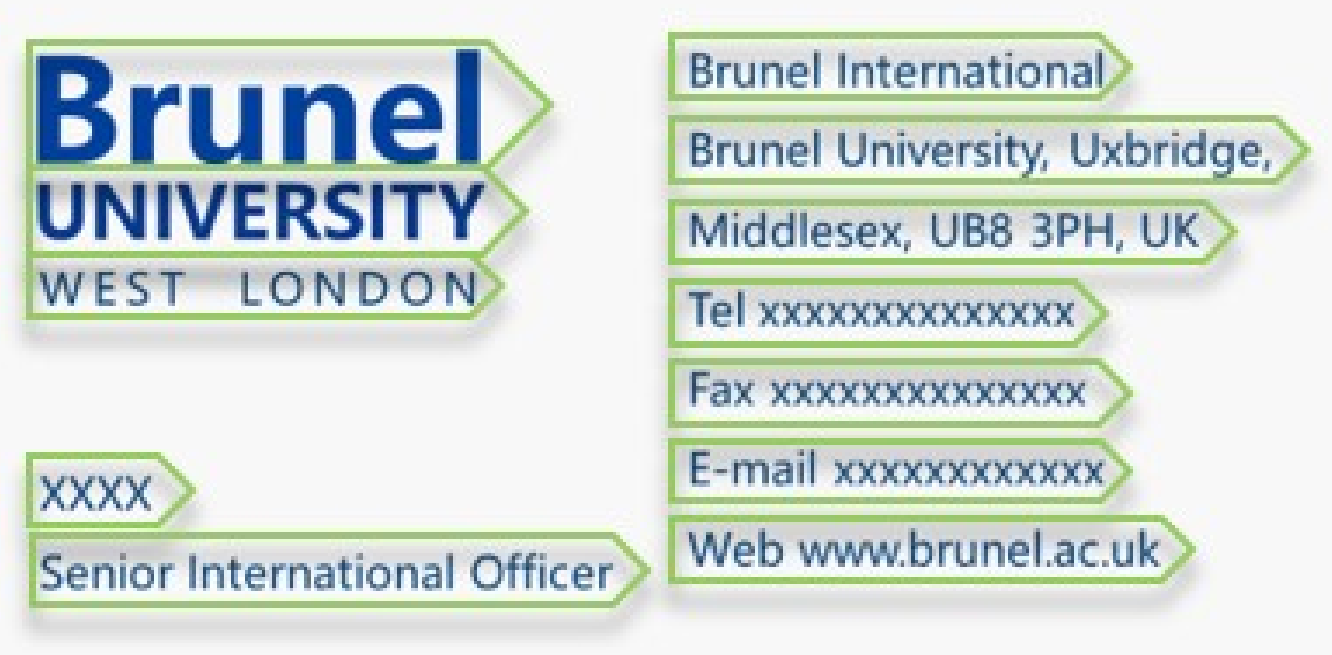}
		}\\
		\subfloat[]{
			\includegraphics[width=0.4\linewidth]{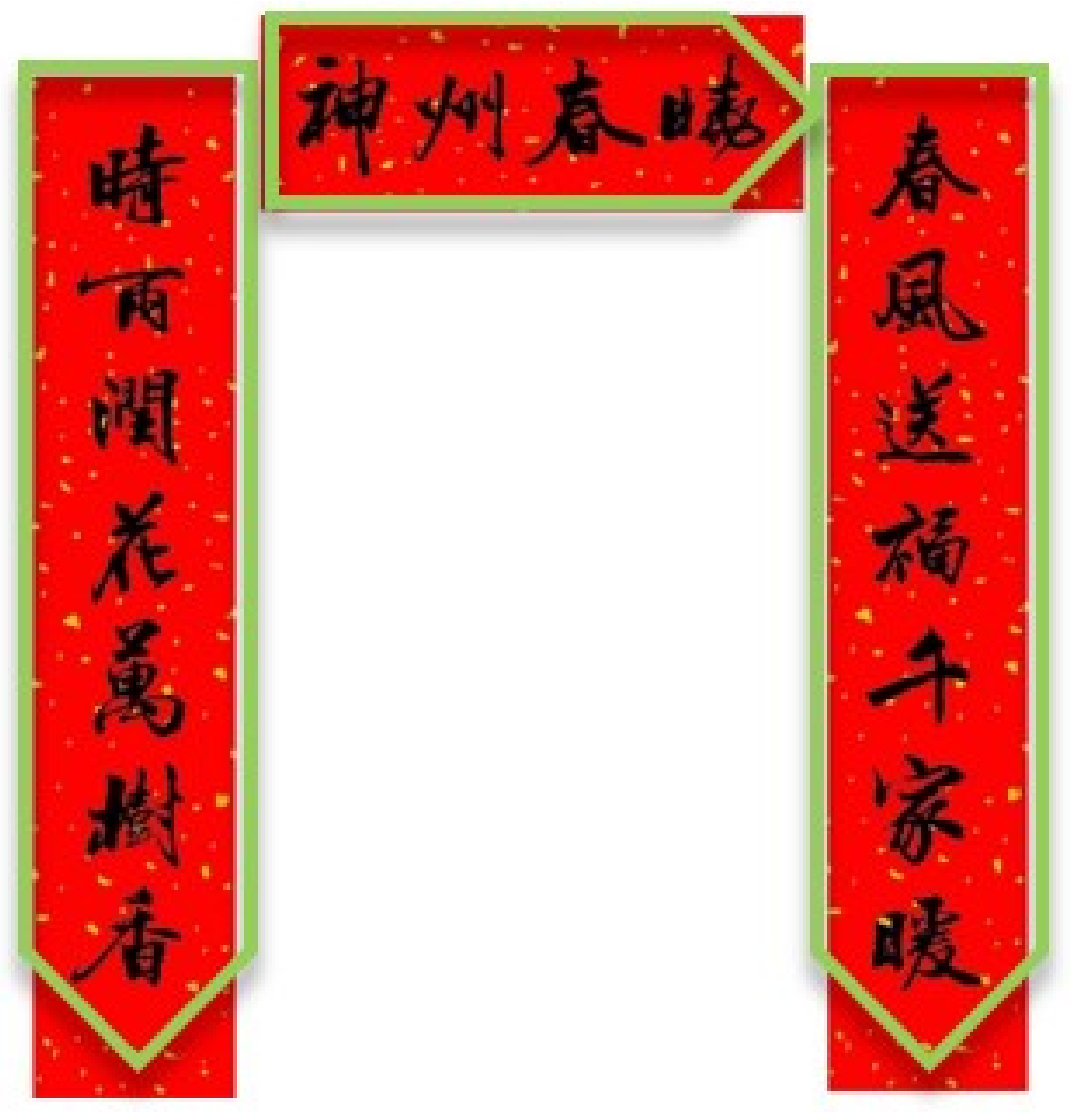}
		}
		\subfloat[]{
			\includegraphics[width=0.55\linewidth]{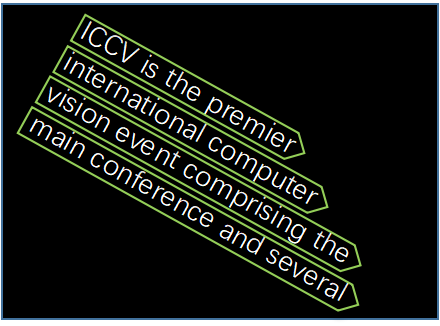}
		}
	\end{center}
	\caption{Examples of the MSR problem.
		(a)-(d) are 4 kinds of multi-sequence scenes.
		Each sequence is bounded by a green box with the arrow indicating text orientation.
		%, and the arrow direction is the internal order of the sequence.
		%(e)-(h) represents relevant examples, where e is an simple image containing two lines text sequence.
		%The one of the sequence is \textsl{helloworld}, another one is \textsl{peace}.
	}	
	\label{fig:unorder_image}
\end{figure}

In this paper, we try to tackle a new and more challenging problem: recognizing multiple text sequences from an image by pure end-to-end learning.
This problem is twofold:
1) \emph{Multiple text sequence recognition}.
Each image may contain multiple \emph{independent} text sequences with different layouts, and we try to recognize all these text sequences. In one word, our problem is a \emph{multiple-sequence recognition} (MSR) problem. We show some MSR examples in Fig.~\ref{fig:unorder_image}.
2) \emph{Pure end-to-end (PEE) learning}.
Each training image is annotated with \textit{only} text transcripts. Our goal is to solve the MSR problem with a PEE method.

Generally, existing NEE and QEE methods cannot handle our problem, because they are not PEE methods. The method of Bluche \etal~\cite{bluche2017scan} can neither. Though it is a PEE method, it is for the SSR problem, at least cannot be straightforwardly applied to our problem. Therefore, we have to explore new method. For better understanding the problem we try to address and the differences between our work and the existing works, we compare NEE, QEE and PEE in Tab.~\ref{tab:CNQP}, and highlight the differences between our work and the typical existing works in Tab.~\ref{tab:works}.

\begin{table}[htb]
	\begin{center}		\caption{Comparison of NEE, QEE and PEE. `T' means text transcripts and `G' means geometric annotations; `D' means detector and `R'means recognizer.}
		\newsavebox{\tablebox}
		\begin{lrbox}{\tablebox}
			\begin{tabular}{|c|c|c|}
				\hline
				\textbf{Method} & \textbf{Architecture} & \textbf{Annotations} \\
				\hline
				NEE & Separate D/R & T+G \\
				%NEE & Detector+Recognizer & T+G \\
				\hline
				QEE & Joint D-R & T+G \\
				%QEE & Bulky Deep network & T+G \\
				\hline
				PEE & R & T \\
				%PEE & Smart Deep network & T \\
				\hline
			\end{tabular}
		\end{lrbox}
		\scalebox{1}{\usebox{\tablebox}}
		\label{tab:CNQP}
	\end{center}
\end{table}
\begin{table}[htb]
	\begin{center}\caption{A comparison of our work with typical existing works from two perspectives: \textit{addressed problem} (MSR or SSR) and \textit{employed method} (NEE, QEE or PEE).}
		\newsavebox{\tableworks}
		\begin{lrbox}{\tableworks}
			\begin{tabular}{|c|c|c|}
				\hline
				\textbf{Problem} & \textbf{Method} & \textbf{Typical works} \\
				\hline
				MSR & NEE & \cite{liao2018textboxes++,liao2017textboxes,wang2011end,wang2010word}\\ \hline
				MSR & QEE & \cite{he2018end,li2017towards,liu2018fots,lyu2018mask}\\ \hline
				SSR & PEE & \cite{bluche2017scan}\\ \hline
				MSR & PEE & Ours\\
				%        Wang \etal \cite{wang2010word,wang2011end}      & MSR & NEE \\
				%Wang \etal \cite{wang2011end}       & MSR & NEE \\
				%		Liao \etal \cite{liao2017textboxes} & MSR & NEE \\ \hline
				%       Li \etal \cite{li2017towards}       & MSR & QEE \\
				%      Liu \etal \cite{liu2018fots}        & MSR & QEE \\\hline
				%	Bluche \etal \cite{bluche2017scan}  & SSR & PEE \\ \hline
				%	Ours                                & MSR & PEE \\
				\hline
			\end{tabular}
			%he2018end
		\end{lrbox}
		\scalebox{1}{\usebox{\tableworks}}
		\label{tab:works}
	\end{center}

\end{table}

To solve this problem, we first consider the connectionist temporal classification~\cite{graves2006connectionist} (CTC) mechanism and the \emph{attention} mechanism~\cite{bahdanau2014neural}.
Both of them are popular in SSR but cannot directly read images with multiple text sequences well.
Concretely, CTC can map a one-dimensional sequence of character probability distribution to a target sequence, but fails to map multi-dimensional probability distribution (\eg the outputs of convolutional neural network) to multiple target sequences.
Though \emph{attention} can capture the attending area of each character from multi-dimensional feature maps to read multi-line text sequences (\eg Fig.~\ref{fig:unorder_image} (a)), it requires that the order of all lines constituting the sequence be pre-defined (as in~\cite{bluche2017scan}).
In fact, it is usually impractical in some real cases (\eg Fig.~\ref{fig:unorder_image} (b) and (c)).

% inspired
%Abstractly, the {\color{red}TODO...}
Therefore, we propose a novel \textbf{M}ultiple \textbf{S}equence \textbf{R}ecognition \textbf{A}pproach (\emph{abbr}. MSRA) to simultaneously learn multiple text sequences from a multi-dimensional feature maps (\eg the outputs of CNN or MDRNN~\cite{graves2009offline}).
Inspired by the concept of one-dimensional probability path in CTC, MSRA is responsible for selecting the optimal probability paths from a given two-dimensional probability space.
And the path selection operation is the process of finding potential target sequences.
Note that, MSRA is trained from the unordered set of multiple independent text sequences, which means that any ordered sequence-level annotations is acceptable during training. % rather than requiring a fixed order among all characters.
%That is, each text sequence is treated as a instance, and we define multiple text sequences as a text instance set, {\color{red}XXX} is trained from the instance set.
For example, annotations in Fig.~\ref{fig:unorder_image}(a) can be interpreted as \{``helloworld'', ``peace''\} or \{``peace'', ``helloworld''\}.

Major contributions of this paper are as follows:
\begin{enumerate}
	\item Conceptually, we propose a new taxonomy of text recognition methods, i.e., NEE, QEE and PEE, and subsume the existing text sequence recognition works into two types: single sequence recognition (SSR) and multiple sequence recognition (MSR). Then, we put forward a new and more challenging problem: recognizing multiple sequences from images by pure end-to-end learning, i.e., MSR by PEE.%  generating multiple text sequences from low-complexity of images.
	
	\item We develop a novel PEE method MSRA to solve the MSR problem, in which the model is trained with only sequence-level text transcripts.
	
	%我们提出了一种面向多序列的方法 without 序列分割与位置标定
	\item As we address a new problem, for evaluating the proposed method, we build up several datasets mainly based on the MNIST dataset and some real application scenarios including automatic bank card reading and ID card reading. %, and achieve promising results comparing to the controlled experiments. % attention based used the \textbf{CTC}, \textbf{ATTENTION}, and \textbf{POL} methods to conduct experiments on multi-line mnist data sets to evaluate the performance of method.
	%2)我们分别使用CTC、ATTENTION、POL方法在多行mnist数据集上进行了实验来评估提出的方法。
	
	\item We conduct extensive experiments on these datasets, which show that the proposed method can effectively recognize multiple sequences from images, and outperforms two CTC/attention based baseline methods. %For further evaluating the effectiveness of our method, we design and release a set of useful synthesis datasets (bank card, identification card and a {\color{red}xxx}) by referring some real-life demands, and test our method on these datasets.% The performance on such datasets also is evaluated by our method.
\end{enumerate}
%introduce a multi-line handwritten digits dataset consisting of 180000 images contain one to six lines of up to 15 character sequences that each character in it is rotated and dithered which is from MNIST dataset.We also introduce datasets in some common scenarios. Bank Card dataset consisting of 100000 images contain 3 lines text.  ID Card dataset consisting of 100000 images contain 2 lines text.

%{图1：a两行的文本信息 }
%-------------------------------------------------------------------------
%创新点：1.多个序列一起训练识别  2.一种不同于以前的端到端

%------------------------------------------------------------------------
\section{Related Work}
Here, we review the major related works based on a new taxonomy of the existing solutions, which are subsumed to three types: NEE, QEE and PEE methods.

\textbf{NEE}.
Wang \etal~\cite{wang2011end,wang2010word} detected and recognized each character in an image with a sliding window, where the word is generated based on characters' global layout.
However, the performance of such methods is limited due to poor representation of handcrafted features.
After that, a number of approaches were proposed for solving the text localization task or the cropped text recognition task better.

\emph{For text localization}, some recent methods~\cite{hu2017wordsup,jiang2017r2cnn,liao2018rotation,liu2017deep,wang2018geometry,zhong2018anchor,zhou2017east} devoted to localizing entire text region with deep neural network techniques, instead of localizing characters one-by-one. These methods were trained with geometric annotations and greatly got better text region detection performance.

\emph{For text recognition}, some early works designed complex handcrafted features such as connected components~\cite{neumann2012real} or Hough voting~\cite{yao2014strokelets} for character recognition, then integrated characters into words.
Later, Jaderberg \etal~\cite{jaderberg2016reading} conducted a 90k-class classification task with a CNN, and generated words with a structured output layer, where each word corresponds to a class. But performance of such methods is impacted by the number of classes, and they cannot be extended to general character sequence generation.
Recent works tended to treat the characters generation process as a sequence learning problem.
Some CTC-based~\cite{shi2017end,wang2017gated} and attention-based~\cite{bai2018ep,cheng2017focusing,cheng2018aon,lee2016recursive,shi2016robust,shi2018aster} methods were proposed for generating character sequence from a cropped text image, which were trained with text transcripts and achieved a big stride forward in performance.

Recently, Liao \etal \cite{liao2018textboxes++,liao2017textboxes} integrated a text region detector and a text recognizer for end-to-end spotting text.
% use an SSD [36] based method for text detection and CRNN [44] for text recognition.

\textbf{QEE}.
The success of deep learning in computer vision inspires a technical wave of reading text from images in an ``end-to-end'' way.
Pioneering works in this direction~\cite{he2018end,li2017towards,liu2018fots} spotted text from images by first pre-training the detection module or recognition module, and then conducting a jointly training on the detection and recognition tasks.
Then, Lyu \etal~\cite{lyu2018mask} proposed an end-to-end trainable text spotter by employing the character-level geometric annoatations.
Such methods still need both text transcripts and geometric annotations to train the model, so they are called \emph{quasi-end-to-end} (QEE) methods.

\textbf{PEE}.
Very recently, Bluche \etal~\cite{bluche2017scan} proposed a method to read multi-line text with a well-designed attention model, in which the order of all text lines constituting the sequence must be pre-defined. So what they handled is actually a SSR problem, only the text sequence may be split to several lines with predefined order and similar orientation.

In this paper, we address a new and more challenging problem: recognizing multiple text sequences from an image with only text (but no location) annotations. Here, the text sequences in the image are independent, they may have different text orientations, and no spatial order constraint exists among them. We propose a PEE method MSRA to solve this problem.
Inspired by the idea of HMM-RNN hybrid~\cite{bengio1999markovian, bourlard2012connectionist}, MSRA directly maps the multi-dimensional outputs of network to all possible target sequences.

%------------------------------------------------------------------------
\section{The MSRA Method}
Here, we present the MSRA method in detail, including the problem formulation, algorithms, loss function and the training  of MSRA as well as prediction using MSRA.
%-------------------------------------------------------------------------
\subsection{Formulation}
MSRA aims to transform a multi-dimensional distribution $\mathbf{X} \in \mathbb{R}^{H\times W \times Q}$ (\eg the forward results of CNN or MDRNN) as a conditional probability distribution over multiple character sequences, where $Q$, $H$ and $W$ are the number of character classes, the height and width of feature maps, respectively.
Formally, $\mathbf{X}$ is represented as
%{\color{yellow}{We need to tell people X is also called \emph{two-dimensional probability distribution.}}}
\begin{equation}
\mathbf{X}= \left( \begin{array}{cccc}
x^{00} & x^{01} & \ldots & x^{0W^\prime} \\
x^{10} & x^{11} & \ldots & x^{1W^\prime} \\
\vdots & \vdots & \ddots & \vdots \\
x^{H^\prime 0} & x^{H^\prime 1} & \ldots & x^{H^\prime W^\prime} \\
\end{array} \right),
\label{eq:softmax}
\end{equation}
where $H^\prime$=$H$-1 and $W^\prime$=$W$-1, and $x^{i,j}\in \mathbb{R}^{Q}$ means the probability distribution at location $(i,j)$. %, and $x^{i,j}$ means the probability of observing label k at time $(i,j)$.

A
%{\color{red}{A}}
character sequence is defined as a sequence of characters $\mathbf{l} \in (\Sigma^*)^*, |\mathbf{l}| > 0 $, in which $\Sigma^*$ is the set of all characters and an extra symbol \textsl{blank} representing character interval or non-text area.
Then, the target 
$\mathbf{Z} = \{\mathbf{l}_1, \mathbf{l}_2, ..., \mathbf{l}_N\}$
%{\color{red}{$\mathbf{Z} = \{\mathbf{l}_1, \mathbf{l}_2, ..., \mathbf{l}_N\}$}} 
is denoted as a set of multiple text sequences.
Hereby, MSRA is devoted to maximizing the conditional probability
%There are \textit{n} sequences $\mathbf{l}_i$ in the target $\mathbf{Z}$, and each of them is at the target price as we show at \ref{te:target_price}.
\begin{equation}
\textit{p}(\mathbf{Z}|\mathbf{X})\overset{def}{=}\frac{1}{N}\sum\limits_{i=1}^N \textit{p}(\mathbf{l}_i|\mathbf{X}),
\label{eq:pzx}
\end{equation}
where $N$ is the number of sequences in $\mathbf{Z}$.
Because of the independence between any two different character sequences, the calculation of $\textit{p}(\mathbf{Z}|\mathbf{X})$ is equivalent to maximizing the conditional probability $\textit{p}(\mathbf{l}|\mathbf{X})$ of each sequence over the input $\mathbf{X}$.
Algorithms for evaluating $\textit{p}(\mathbf{l}|\mathbf{X})$ are presented in the following subsection.

%-------------------------------------------------------------------------
\subsection{Algorithms}\label{sec:3.2}
\begin{figure}[t]
	\begin{center}
		\includegraphics[width=1\linewidth]{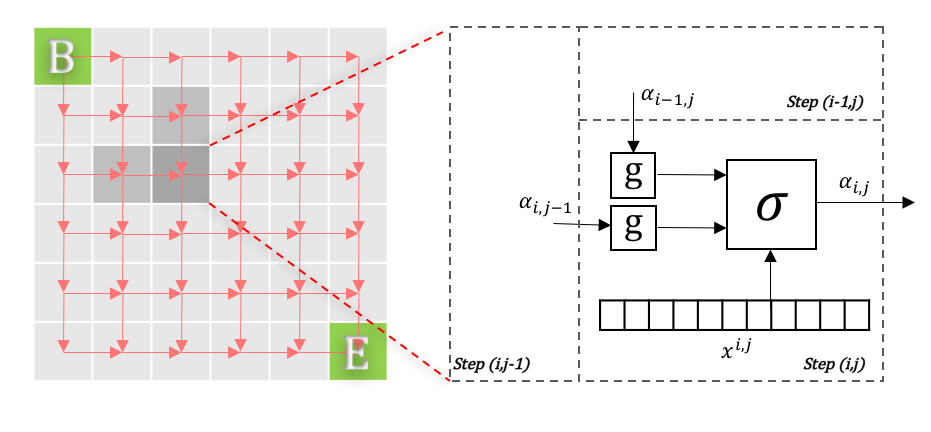}
	\end{center}
	\caption{
		The illustration of path generation on the two-dimensional probability {{distribution}} $\mathbf{X}$.
		$B$ and $E$ separately denote the beginning position and ending position of path searching.
		For a cell $(i,j)$ in $\mathbf{X}$, its state value $\alpha_{i,j}$ depends on both $\alpha_{i,j-1}$ and $\alpha_{i-1,j}$. %, which is determined with a control parameter $\sigma$.%, that is, $\alpha^{i,j}=\sigma(g(\alpha^{i,j-1}),g(\alpha^{i-1,j}))$. %After that, $\phi^{i,j} = ???$. $\sigma_1$ and $\sigma_2$ are two linear functions.
		%There are begin step\textit{(B)} and end step\textit{(E)} in the network output $\Gamma$, and there are multiple state transition paths from B to E.
		%Each step can be state transitioned with the adjacent step of any dimension.
		%There is only one state transition in the step of the feature map's boundary $\textit{(the time of any dimension is 0)}$, except that the other steps have two states transitions.
		%In the dark gray step $(i,j)$ of $\Gamma$, the selection gate $d=\sigma_1(\phi^{i,j-1},\phi^{i-1,j})$ is a linear function processing input from adjacent steps.
		%The linear combination of the two inputs is passed along with the probability distribution column $x^{i,j}$ to the linear function $\sigma_2$ to calculate the step's output $\phi^{i,j}$.
		%在网络输出$\Gamma$中有起始时间B和终止时间E，B到E存在多条状态转移路径。
		%每个时间状态可以与任一维度的相邻时间状态进行状态转移，在时间序列边界（任一维度的时间为0）的时间状态仅有一种状态转移，除此外的时间状态均有两种状态转移。
		%例如在$\Gamma$中的深灰色时间状态$(x,y)$，选择门$\mathcal{d}=\sigma_1(\phi^{x,y-1},\phi^{x-1,y})$是一个线性函数，将两个输入的线性组合与概率分布列$\gamma^{x,y}$一起传递给线性函数$\sigma_2$计算得到其输出$\phi^{x,y}$。
	}
	\label{fig:softmaxout}
\end{figure}

$\textit{p}(\mathbf{l}|\mathbf{X})$ is the mapping condition probability of $\mathbf{l}\in \mathbf{Z}$ over the learnt two-dimensional {{probability distribution}} $\mathbf{X}$.
By extending the concept of one-dimensional probability path in CTC, the evaluation of $\textit{p}(\mathbf{l}|\mathbf{X})$ turns to solve the three-dimensioned probability path $\bar{l}$ search problem over $\mathbf{X}$, where $\bar{l}$ is a path (as shown in Fig.~\ref{fig:softmaxout}) from the beginning $x^{0,0}$ to the end $x^{H^\prime,W^\prime}$. $\bar{l}$ is then mapped to a label sequence $\mathbf{l}$ by using the \emph{many-to-one} $\mathcal{B}$-mapping strategy in CTC~\cite{graves2006connectionist}.
Therefore, $\textit{p}(\mathbf{l}|\mathbf{X})$ can be further rewritten as
\begin{equation}
\textit{p}(\mathbf{l}|\mathbf{X})=\sum\limits_{\bar{l} \in \mathcal{B}^{-1}(\mathbf{l})} \textit{p}(\bar{l}|\mathbf{X}).
\label{eq:l_to_lbar}
\end{equation}
In Eq.~(\ref{eq:l_to_lbar}), $\textit{p}(\bar{l}|\mathbf{X})$ is further transferred as the prefix sub-path search problem, which can be iteratively calculated with a dynamic programming algorithm, as shown in the \emph{Forward Algorithm}.

For representing the non-text areas, we also adopt: 1) the label sequence extension strategy, that is, adding blanks to the beginning and the end, and inserting blanks between each pair of neighboring characters, to get an extended sequence $\mathbf{l^\prime}$; and 2) the state transfer strategy, i.e., allowing path transitions between blank and any non-blank character, and any pair of distinct non-blank characters.

\subsubsection{Forward Algorithm}

\begin{figure}[t]
	\begin{center}
		\subfloat[]{
			\includegraphics[width=0.45\linewidth]{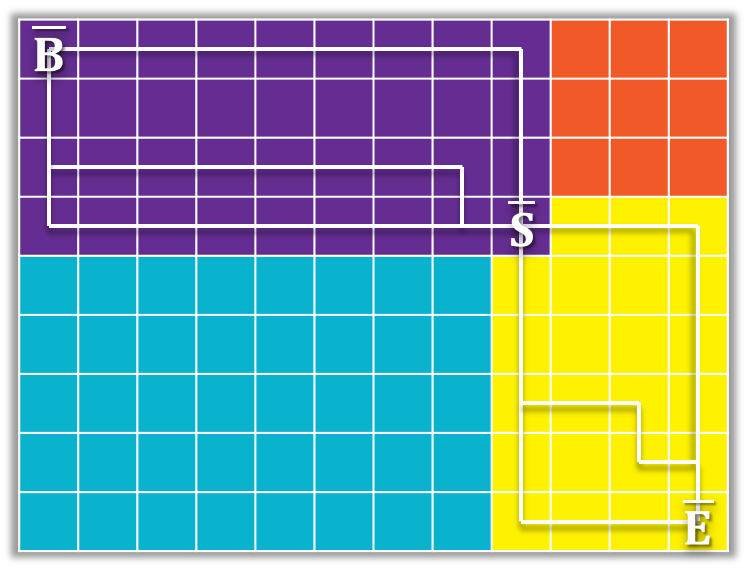}
		}
		\subfloat[]{
			\includegraphics[width=0.45\linewidth]{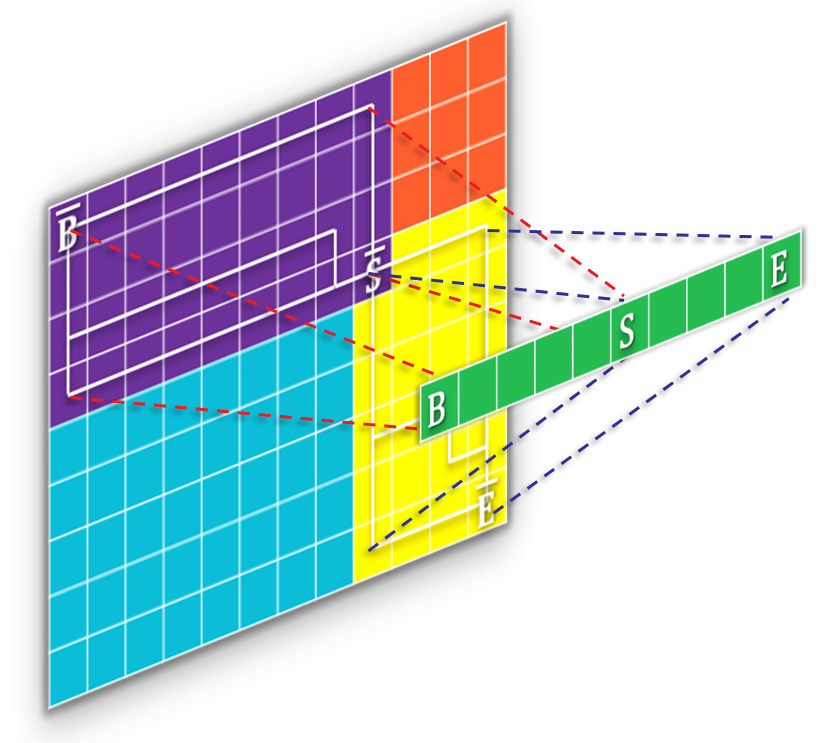}
		}
	\end{center}
	\caption{
		The illustration of the forward and backward algorithms matching the $s$ position of $\mathbf{l}^\prime$ at $\bar{S}(i,j)$.
		In subgraph (a), the dark purple area represents the path search area of the forward algorithm, where the white paths $\bar{l}$ from $\bar{B}$ to $\bar{S}$ are all solutions satisfying $\mathcal{B}(\bar{l})=\mathbf{l}^\prime_{0:s}$.
		The yellow area represents the path search area of the backward algorithm, where the white paths $\bar{l}$ from $\bar{S}$ to $\bar{E}$ are all solutions satisfying $\mathcal{B}(\bar{l})=\mathbf{l}^\prime_{s:|\mathbf{l}^\prime|-1}$.
		%According to the state transition of Fig.\ref{fig:softmaxout}, $\mathbf{l}^\prime$ does not pass through the blue and orange areas, so they are called the irrelevant area.
		In subgraph (b), the green sequence represents $\mathbf{l}^\prime$, in which $B$,$S$ and $E$ correspond to the location 0,s and $|\mathbf{l}^\prime|-1$, respectively.
		The forward calculation maps the paths from $\bar{B}$ to $\bar{S}$ in the deep purple area to the sequence from $B$ to $S$, and the backward calculation maps the paths from $\bar{S}$ to $\bar{E}$ in the yellow area to the sequence from $S$ to $E$.
	}
	\label{fig:path}
\end{figure}

We define $\alpha_{i,j}(s)$ as the probability for 
%the prefix of 
$\bar{l}$ matching $\mathbf{l}^\prime_{0:s}$ at $(i,j)$
\begin{equation}
\begin{split}
\alpha_{i,j}(s) \overset{def}{=} \sum\limits_{\substack{\bar{l} \\ \mathcal{B}(\bar{l})=\mathbf{l}^\prime_{0:s}}}
\prod\limits_{t=0}^{|\bar{l}|-1} x_{\bar{l}_t}^{i_t,j_t},  \\
\end{split}
\label{eq:ad}
\end{equation}
%{\color{red}what is $i_t$, $j_t$ and $l_t$?????},
where $\bar{l}_t$ is the traversal of the path $\bar{l}$, and $i_t$, $j_t$ are the coordinates of the step that matches $\bar{l}_t$.
In Fig.~\ref{fig:path}(a), the white paths in the dark purple area (in the top-left corner) illustrate Eq.~(\ref{eq:ad}). Concretely, each of these white paths corresponds to an $\bar{l}$ in Eq.~(\ref{eq:ad}).
%$\alpha_{i,j}(s)$ is shown as the white path from $\bar{B}$ to $\bar{S}$ in Figure. \ref{fig:path}.
%The white path of the dark purple area in Fig.\ref{fig:path} is the visualization of Eq.\ref{eq:ad}.
%Fig.\ref{fig:path}深紫色区域的白色路径就是Eq.\ref{eq:ad}的可视化结果。

The initialization rules of Eq.~(\ref{eq:ad}) are as follows:
\begin{equation}
\begin{split}
&\alpha_{0,0}(0) = x_b^{0,0}, \\
&\alpha_{0,0}(1) = x_{\mathbf{l}_1}^{0,0}, \\
&\alpha_{0,0}(s) = 0, \forall s > 2.
\end{split}
\end{equation}
$\alpha_{i,j}(s)$ can be iteratively calculated by
%The prefix-path matching at $(i,j)$ can be calculated based on its prefix path matching
\begin{equation}
\begin{split}
%\alpha_{i,j}(s) = (d_1 g_{i-1,j}(s) + d_2 g_{i,j-1}(s))x_{\mathbf{l}^\prime_s}^{i,j},
\alpha_{i,j}(s)=\sigma(g(\alpha_{i,j-1}, s),g(\alpha_{i-1,j}, s)) \\
=\lambda_1 g(\alpha_{i,j-1}, s) + \lambda_2 g(\alpha_{i-1,j},s)
\label{eq:alpha_d}
\end{split}
\end{equation}
where $\sigma$ is a linear function for evaluating the path transitions from $(i-1, j)$ or $(i, j-1)$. 
$\lambda_1$, $\lambda_2$ are the hyper-parameters of $\sigma$. They will be detailed in the \emph{Performance Evaluation} section.

And $g(\alpha_{i,j}, s)$ is denoted as
\begin{equation}
\begin{split}
& g(\alpha_{i,j}, s) \overset{def}{=} (\alpha_{i,j}(s) + \alpha_{i,j}(s-1) + \eta\alpha_{i,j}(s-2)) x^{i,j}_{l\prime_s}, \\
& \eta =
\begin{cases}
0 & \text{if } \mathbf{l}^\prime_s = \text{blank or } \mathbf{l}^\prime_s = \mathbf{l}^\prime_{s-2},\\
1 & \text{otherwise}.
\end{cases}
\end{split}
\end{equation}

In Eq.~(\ref{eq:alpha_d}), there are two boundary problems to be considered:
\begin{enumerate}
	\item $\alpha_{i,j}(s)=0, \quad \forall s < |\mathbf{l}^\prime| \text{-} 2(H\text{-}i\text{+}W\text{-}j\text{-}1) $ means that there are not enough locations to match the remaining characters in $\mathbf{l}^\prime$.
	%这些状态表示没有足够的位置来匹配剩余\mathbf{l}^\prime中的字符。
	\item $\alpha_{i,j}(s)=0, \quad \forall s > 2(i\text{+}j)\text{+}1$  means that the matching position $s$ is beyond the maximum length currently matchable.
	%这些状态表示匹配位置s超出了目前所能匹配的最大长度。
\end{enumerate}

At last, $\textit{p}(\mathbf{l}|\mathbf{X})$ is the sum of the prefix-path probabilities at $(H^\prime, W^\prime)$, i.e.,
\begin{equation}
\textit{p}(\mathbf{l}|\mathbf{X})=\alpha_{H^\prime,W^\prime}(|\mathbf{l}^\prime|-1) +\alpha_{H^\prime,W^\prime}(|\mathbf{l}^\prime|-2).
\label{eq:calcbyalpha}
\end{equation}

\subsubsection{Backward Algorithm}

Similarly, we define $\beta_{i,j}(s)$ as the probability for %the suffix of 
$\bar{l}$ matching $\mathbf{l}^\prime_{s:|\mathbf{l}^\prime|-1}$ at $(i,j)$ but not relying on $x_{\bar{l}_{0}}^{i_0,j_0}$, that is,
\begin{equation}
\begin{split}
\beta_{i,j}(s) \overset{def}{=} \sum\limits_{\substack{\bar{l} \\ \mathcal{B}(\bar{l})=\mathbf{l}^\prime_{s:|\mathbf{l}^\prime|-1}}}
\prod\limits_{t=1}^{|\bar{l}|-1} x_{\bar{l}_t}^{i_t,j_t}, \\
\end{split}
\label{eq:bd}
\end{equation}
where $\bar{l}_t$ is the traversal of the path $\bar{l}$, and $i_t$, $j_t$ are the coordinates of the step that matches $\bar{l}_t$.
%$\beta_{i,j}(s)$ is shown as the white path from $\bar{S}$ to $\bar{E}$ in Figure. \ref{fig:path}.
In Fig.~\ref{fig:path}(a), the white paths in the yellow area (in the bottom-right corner) illustrate Eq.~(\ref{eq:bd}). Concretely, each of these white paths corresponds to an $\bar{l}$ in Eq.~(\ref{eq:bd}).

%Fig.\ref{fig:path}黄色区域的白色路径就是Eq.\ref{eq:bd}的可视化结果。
The initialisation rules for backward algorithm are as follows:
\begin{equation}
\begin{split}
&\beta_{H^\prime,W^\prime}(|\mathbf{l}^\prime|-1) = 1, \\
&\beta_{H^\prime,W^\prime}(|\mathbf{l}^\prime|-2) = 1, \\
&\beta_{H^\prime,W^\prime}(s) = 0, \forall s < |\mathbf{l}^\prime|-2.
\end{split}
\end{equation}
Thus, $\beta_{i,j}(s)$ can be iteratively calculated as
\begin{equation}
\beta_{i,j}(s) = \lambda_1 g^\prime(\beta_{i,j+1}, s) + \lambda_2 g^\prime(\beta_{i+1,j}, s),
\label{eq:beta_d}
\end{equation}
where
\begin{equation}
\begin{split}
& g^\prime(\beta_{i,j}, s) \overset{def}{=} \beta_{i,j}(s)x_{\mathbf{l}^\prime_s}^{i,j} \text{+} \beta_{i,j}(s\text{+}1)x_{\mathbf{l}^\prime_{s\text{+}1}}^{i,j} \text{+} \eta^\prime\beta_{i,j}(s\text{+}2)x_{\mathbf{l}^\prime_{s\text{+}2}}^{i,j}, \\
& \eta^\prime =
\begin{cases}
0 & \text{if } \mathbf{l}^\prime_s = \text{blank or } \mathbf{l}^\prime_s = \mathbf{l}^\prime_{s+2},\\
1 & \text{otherwise}.
\end{cases}
\end{split}
\end{equation}
The boundary considerations are the same as that in $\alpha$ computation.

%Intuitive we can calculate the label sequence probability $\textit{p}(\mathbf{l}|\mathbf{X})$ by Eq.\ref{eq:calcbyalpha}.
%Fig.\ref{fig:path} provides another idea. Through (b) we can more intuitively see how the network output $\Gamma$ is mapped to the sequence $\mathbf{l}^\prime$.
%In this case, the product of the forward variable $\alpha$ and the backward variable $\beta$ of $\bar{S}$ represents the feasible probability of the complete matching sequence $\mathbf{l}^\prime$.
%That is to say, the label sequence probability can also be given by the sum of the forward and backward variables at any position.

%\begin{equation}
%\textit{p}(\mathbf{l}|\mathbf{X})= \sum\limits_{s=0}^{|\mathbf{l}^\prime|} \alpha_{s}^{x,y}\beta_{s}^{x,y}
%\end{equation}

%-------------------------------------------------------------------------
\subsection{Objective Function and Model Training}
The objective function of MSRA is defined as the negative log probability of the correct complete-matching input set $\mathcal{S}$, that is
\begin{equation}
O = - \sum\limits_{(\mathbf{X},\mathbf{Z}) \in \mathcal{S} } \ln\textit{p}(\mathbf{Z}|\mathbf{X}).
\label{eq:objective}
\end{equation}

In order to train the model with the standard back-propagation algorithm~\cite{rumelhart1986learning}, we need to derive the partial derivative of the objective function \textsl{O} with regards to $x_k^{i,j}$.
By substituting Eq.~(\ref{eq:pzx}) into Eq.~(\ref{eq:objective}), we can get
\begin{equation}
O = - \sum\limits_{(\mathbf{X},\mathbf{Z}) \in \mathcal{S} } (\ln\sum\limits_{i=1}^N \textit{p}(\mathbf{l}_i|\mathbf{X})-\ln N).
\label{eq:o_lx}
\end{equation}

The gradient solving problem is transformed from a multi-sequence recognition problem to a single-sequence recognition problem.
Then, we have
\begin{equation}
\frac{\partial \textit{p}(\mathbf{l}|\mathbf{X})}{\partial x_k^{i,j}} = \frac{1}{x_k^{i,j}} \sum\limits_{s \in lab(\mathbf{l},k)} \alpha_{i,j}(s)\beta_{i,j}(s)
\label{eq:lx_y}
\end{equation}
where $lab(\mathbf{l},k)={\{ s:\mathbf{l}^\prime_s=k \} }$.
Finally, based on Eq.~(\ref{eq:o_lx}) and Eq.~(\ref{eq:lx_y}), we have
\begin{equation}
\frac{\partial O}{\partial x_k^{i,j}} = - \frac{1}{x_k^{i,j}\sum\limits_{t=1}^n \textit{p}(\mathbf{l}_t|\mathbf{X})} \sum\limits_{t=1}^n \sum\limits_{s \in lab(\mathbf{l}_t,k)} \alpha_{i,j}(s)\beta_{i,j}(s).
\end{equation}

%\subsection{POL training process}
%We use two feature extractors, one based on VGG and one based on ResNet.
%Their performance evaluation will be given in 4.x\cite{rumelhart1986learning}.
%During training, we can set the hyperparameter of the selection gate according to the problem scene.
%For example, when we solve the problem of multi-line text sequence recognition problem, the parameter of the selection gate is set to (0.9, 0.1), so that the training pays more attention to the horizontal text.
%我们使用了两种特征提取器，一种是基于VGG的，一种是基于ResNet的。其性能评估将会在4.x给出。
%在训练时，我们可以根据问题场景设置选择门的超参数，例如在我们解决多行文本序列问题时，选择门的参数设置为0.9,0.1，使训练更关注横向文本。

%-------------------------------------------------------------------------

\subsection{Prediction}
\begin{figure}[t]
	\begin{center}
		\includegraphics[width=0.95\linewidth]{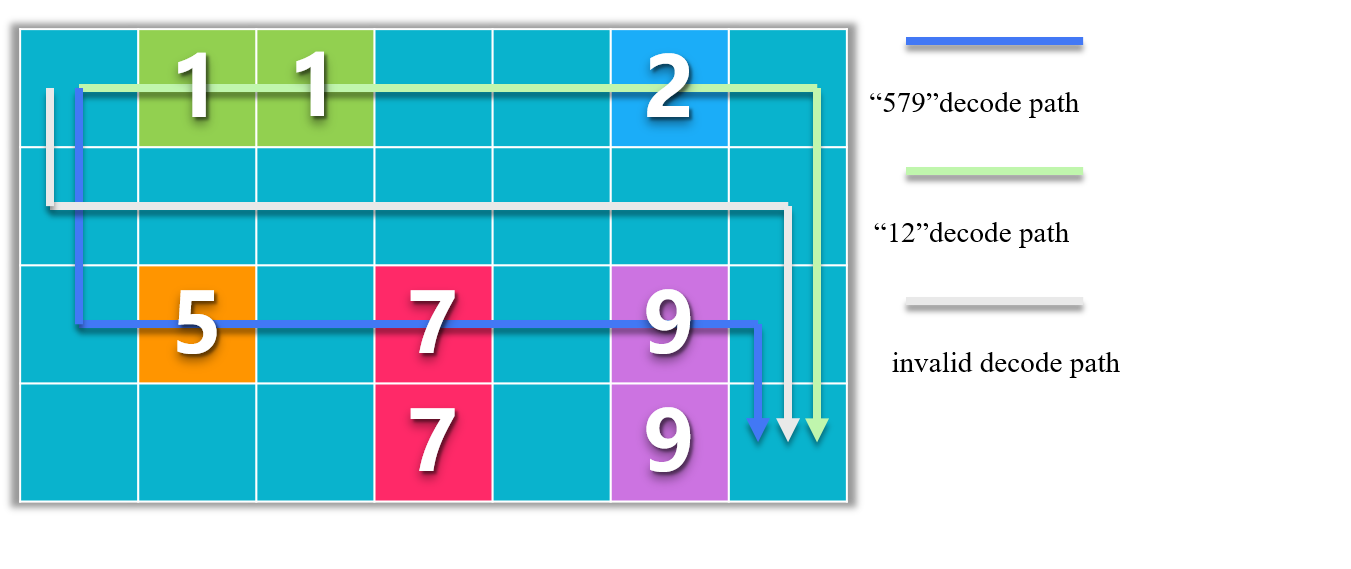}
	\end{center}
	\caption{
		The illustration of the learnt maximum probability matrix of $\mathbf{X}$.
		Numbers in cells indicate the labels of predicted character classes, and blank cells have no character.
		%The maximum probability class of each step is given in terms of color and number.And the class of the step without number is blank.
		The green path and blue path correspond to targets ``12'' and ``579'' respectively, while the gray path has only blanks.
		%图1的最大类别概率图。
		%时间步为绿色表示此处最大概率类别为空白类，蓝色则对应文本类。
		%\{helloworld; peace\}可由红色路径对应的类别序列经过$\mathcal{B}$转换后得到。
		%白色路径展示了一条从起始时间到终止时间的空白路径，在一维时间序列中这是不可能存在的。
	}
	\label{fig:maxclassific}
\end{figure}

%POL is a method based on the encoder-decoder framework.
The prediction process of MSRA is to generate multiple text sequences $\mathbf{Z}$ from the learnt  two-dimensional {{probability}} distribution $\mathbf{X}$.
%The decoding scheme requires us to decode the target $\mathbf{Z}$ from the three-dimensional space (two-dimensional is time, one-dimensional is class probability).
%The decoding direction in one-dimensional space is easy to set, that is, decoding from left to right.
%Even the prefix search decoding method\cite{graves2006connectionist}, which consumes a lot of resources, follows this decoding order.
%POL是基于编码解码框架的方法。
%解码方案要求我们从三维空间(两维时间，一维类别概率)中解码出目标$\mathcal{Z}$。
%在一维空间中解码方向是容易设定的，即从左至右依次解码即可。
%就算是资源消耗极高的Prefix search decoding\cite{graves2006connectionist}，也是按照这种解码顺序。
In our case, it is impractical to adopt the prefix search strategy used in CTC due to the following reasons:
\begin{itemize}
	\item {{Since each step has two choices, prefix search will result in $O(2^{N}K)$ computational complexity, where $K$ is the computational complexity of the prefix search method in CTC.}}
	
	\item It is highly possible that paths contain only blanks (\eg the gray path in Fig.~\ref{fig:maxclassific}), which makes prefix search totaly fail.
	%Because there is a time series of empty categories in a two-dimensional time series, such as the gray path in Fig.\ref{fig:maxclassific}.
	%     It has no practical significance in the forward and backward algorithms, but it is a possible result in accordance with the new method.
	%    If the time step belonging to it is that the probability of the blank class is extremely high, the path probability is higher than the correct decoding path.
\end{itemize}
%In a two-dimensional time series, each time step has two selectable directions.
%If the direction selection is added to the prefix search decoding method, the resource consumption of the new method will be expanded $2^{n-1}$ times at the nth time step.
%Obviously, it is impractical.

%Because there is a time series of empty categories in a two-dimensional time series, such as the gray path in Fig.\ref{fig:maxclassific}.
%It has no practical significance in the forward and backward algorithms, but it is a possible result in accordance with the new method.
%If the time step belonging to it is that the probability of the blank class is extremely high, the path probability is higher than the correct decoding path.
%二维时间序列中，每个时间步均有两个可选择方向。
%若在Prefix search decoding的基础上加上方向的选择，那么在第n个时间步，新方案的资源消耗将是原本的$2^{n-1}$倍。
%显然是不可行的，并且按照这种思想选择出的路径不一定是答案。
%因为在二维时间序列中存在空类别的时间序列，如图4中的白色路径，其在前后向算法中并没有实际意义，却是一条符合新方案的可能结果。
%若其所属的时间步空白类别概率极高，路径概率是高于可正确解码路径的。
\begin{table*}[!htb]
	\begin{center}
		\newsavebox{\tablenqp}
		\begin{lrbox}{\tablenqp}
			\begin{tabular}{|c|c|c|c|c|c|c|c|c|c|c|c|c|c|c|c|}
				\hline
				\multirow{2}{*}{\textbf{Method}} &
				\multicolumn{3}{c|}{\textbf{MS-MNIST[1]}} &
				\multicolumn{3}{c|}{\textbf{MS-MNIST[2]}} &
				\multicolumn{3}{c|}{\textbf{MS-MNIST[3]}} &
				\multicolumn{3}{c|}{\textbf{MS-MNIST[4]}} &
				\multicolumn{3}{c|}{\textbf{MS-MNIST[5]}}\\
				\cline{2-16}
				& \textbf{NED} & \textbf{SA} & \textbf{IA}
				& \textbf{NED} & \textbf{SA} & \textbf{IA}
				& \textbf{NED} & \textbf{SA} & \textbf{IA}
				& \textbf{NED} & \textbf{SA} & \textbf{IA}
				& \textbf{NED} & \textbf{SA} & \textbf{IA}\\
				\hline
				\textbf{MSRA} & 0.65 & 91.23 & 91.23 &
				0.48 & 93.57 & 87.47 &
				0.74 & 90.19 & 73.23 &
				1.21 & 86.35 & 63.20 &
				1.82 & 77.69 & 27.93 \\
				\hline
				\textbf{Attention based method} & 0.90 & 89.03 & 89.03 &
				0.67 & 91.48 & 83.87 &
				1.25 & 87.52 & 67.27 &
				1.35 & 88.55 & 61.80 &
				88.69 & 0 & 0\\
				\hline
				\textbf{CTC based method} & 0.78 & 89.60 & 89.60 &
				- & - & - &
				- & - & - &
				- & - & - &
				- & - & - \\
				\hline
			\end{tabular}
		\end{lrbox}
		\scalebox{0.8}{\usebox{\tablenqp}}
		\caption{Recognition results on MS-MNIST datasets.
			Only when all sequences in an image are recognized accurately, the image is considered being recognized accurately.
		}
		\label{tab:mlmnist}
	\end{center}
\end{table*}

Even so, the prediction of MSRA can be solved in a task-specific way. That is, we can design the corresponding mapping strategies based on the specific scenarios.
%we need to solve the problem of multi-line text recognition, and MSRA has a very good feature:
%\textit{If a text sequence in the graph is labeled in one row, then this text sequence can be completely matched by a $\mathcal{B}$ conversion of a line of the maximum probability class map $M$(formed by the class with the highest probability of each time step).}
%In Fig.\ref{fig:maxclassific} we can visually see this feature.
%所幸我们需要解决的是多行文本识别问题，并且POL有一个很好的特性：
%图中的一个文本序列只要按一行进行标定，那么在最大概率类别图$M$(由每个时间序列取概率最大的类别形成)中，这个文本序列可以由某行经过$\mathcal{B}$转换后完整匹配。
%在图4中我们可以直观的看到这种特性。
Given the learnt $\mathbf{X}$, we define the mapping function $F$ to map the maximum probability matrix $M=argmax(\mathbf{X})$ to the target $\mathbf{Z}$ by
\begin{equation}
\mathbf{Z} = F(argmax(\mathbf{X})).
\end{equation}

Specifically, the function $F(.)$ is formulated as
\begin{equation}
\begin{split}
F(M) \overset{def}{=} \{\mathcal{B}(T_{r_{1,1}} + T_{r_{1,2}} + ...);\\
\mathcal{B}(T_{r_{2,1}} + T_{r_{2,2}} + ...);...\},
\end{split}
\end{equation}
where
\begin{equation}
T_{r_{i,j}} \overset{def}{=} M_{r_{i,j}}[st_{r_{i,j}},en_{r_{i,j}}] = \sum\limits_{k=st_{r_{i,j}}}^{en_{r_{i,j}}} M_{r_{i,j},k}
\end{equation}
where $r_{i,j}$ represents the row coordinate of the $j$-th string that is a substring of the $i$-th sequence to be generated from $\mathbf{Z}$.
$M_{r_{i,j},k}$ is the value at ($r_{i,j}$, $k$) in $M$.
Then, $F$ can be approximately learnt with a subset $M^\prime$ of $M$ and the corresponding subset $\mathbf{Z}^\prime$ of $Z$. %,  in $\mathcal{S}$, and
The loss function is defined as
%Then, $F$ can be approximately learnt from the maximum probability class map $M'$ corresponding to the $\mathcal{S}'$ which is a part of the input set $\mathcal{S}$  to the result $\mathcal{Z}'$ corresponding to $\mathcal{S}'$.
%In the multi-line text recognition problem, $f$ maps each sequence $\mathbf{l}_i$ in $\mathcal{Z}'$ to a row $M'_j$ in $M'$, which is $f(\mathbf{l}_i)=M'_j$.
%Construction the loss function g
%选择输入集$\mathcal{S}$中的一部分$\mathcal{S}'$对应的结果$\mathcal{Z}'$以及$\mathcal{S}'$对应的最大概率类别图$M'$，构建从$\mathcal{Z}'$到$M'$的映射$f$。
%构建从输入集$\mathcal{S}$中的一部分$\mathcal{S}'$对应的结果$\mathcal{Z}'$到$\mathcal{S}'$对应的最大概率类别图$M'$的映射$f$。
%在多行文本识别问题中，$f$将$\mathcal{Z}'$中的每个序列$\mathbf{l}_i$映射到$M'$中的某一行$M'_j$，即$f(\mathbf{l}_i)=M'_j$ 。
%构造损失函数
\begin{equation}
\mathcal{L} = \frac{1}{N}\sum \limits_{i=1}^{N} \min\limits_j NED(\mathbf{l}_i,\mathcal{B}(T_{r_{j,1}} \text{+} T_{r_{j,2}} \text{+} ...))
\end{equation}
where NED is the edit distance normalized by the number of characters in $\mathbf{l}_i$, and
$N$ is the number of sequences contained in the samples of $\mathbf{Z}^\prime$.
%Select the optimal mapping to get the decoding path in the maximum probability class map, and then get the answer through $\mathcal{B}$ conversion, as shown in Fig.\ref{fig:maxclassific}.
%其中EDN是edit distance normalized by the number of characters in the $\mathbf{l}_i$.
%选取最优的映射得到在最大概率类别图中的解码路径，进而通过$\mathcal{B}$转换得到答案,如图4所示。
%我们将样本中的标定串与特征图建立映射，构造损失函数，选取最优映射进而得到解码路径。
%:画FIG1的最大概率类别图，并解释全部走空白 类的可能反驳prefix search变种的不能性，并解释训练完成后最大概率图中每个文本序列均在一行上。

\section{Performance Evaluation}
In this section, we first evaluate the effectiveness of MSRA by comparing it with two CTC-based and attention-based baseline methods on multiple-sequence datasets generated based on MNIST~\cite{lecun1998gradient}, and visualize multiple sequences generation (the maximum probability paths) in $\alpha$ space.
Then, we evaluate our method on four other datasets generated mainly based on two real application scenarios, including automatic bank card reading and ID card reading.

%Images in the popular datasets contain a great deal of geometric information which is difficult to learn without geometric annotations. So we create new datasets with a small amount of geometric information to evaluate MSRA. 

Note that this work addresses a new and very challenging problem,  
we evaluate our method on several synthetic datasets, and compare it with two CTC and attention based baseline methods.
As an innovative and exploring work, we humbly believe that the current evaluation is enough to validate the effectiveness of the proposed method. We intend to leave further evaluation on popular datasets and comparison with other existing methods for a future work.

%version 0321
%Note that this work addresses a new problem, we evaluate our method on several datasets created by ourself specially for this problem, and we do not compare our method with specific existing methods. We only compare our method with two CTC and attention based baseline methods. 

%At last, we visualize the
%Our experiment is divided into two parts.
%In the first part, we evaluated the performance of this method on a multi-sequence dataset based on MNIST synthesis and compared it with sequence-oriented method ctc and attention.
%In the second part, we tested the performance of this method in the bank card and ID card dataset.
\begin{figure}[htp]
	\begin{center}
		\includegraphics[width=0.99\linewidth]{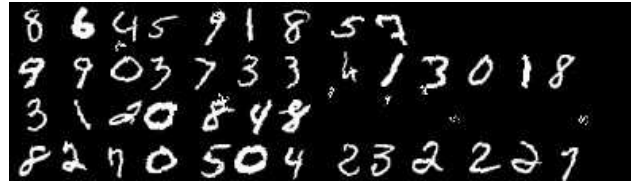}
	\end{center}
	\caption{
		An example of multiple-sequence samples based on MNIST.
	}
	\label{fig:mlmnistl4}
\end{figure}

\subsection{Performance on MNIST based Datasets}
\subsubsection{Datasets}
We create MNIST-based multiple-sequence datasets by randomly selecting some digits from MNIST, and connecting them into a sequence. We superimpose no more than 5 such sequences into an image.
%The dataset is constructed by taking some digits from MNIST and connecting them into a sequence, and then superimposing some sequences.
Concretely, the process is like this: 1) each digit taken from MNIST has a size of 28 $\times$ 28 pixels, a horizontal-offset of $\pm$3 pixels, and a rotation angle of $\pm$10; 2) The number of sequences in images follows normal distribution approximately; 3) Each sequence consists of no more than 14 above-mentioned digits, and sequence length obeys normal distribution approximately; 4) Sequences are superimposed to each image from top to bottom. Correspondingly, if the number of sequences in an image is $k$, the size of the image is 28$k$ $\times$ 392.
In addition, {{we add some noise data to each image. The noise data are digits}}  randomly taken from MNIST with size of 7$\times$7 pixels. The number of noise digits is one fifth of the valid digits contained in the image. Noise digits are superimposed to randomly selected positions in each image.

In such a way, we create five datasets, which are denoted by MS-MNIST[$n$] where $n$ means that each image in the dataset contains at most $n$ sequences, and $n$=1, 2, 3, 4 and 5.
%It is called MLmnistl[n] where n represents the maximum number of sequences in the dataset is n.
%In the dataset, the number of sequences in the image obeys normal distribution approximately.
%The sequence is superimposed from top to bottom.
%Each digit take from MNIST has a size of 28 $\times$ 28 pixels, a horizontal offset of $\pm$3 pixels, and a rotation angle of $\pm$10.
%Each sequence consists of no more than 14 of the above-mentioned digital elements, and the length obeys normal distribution approximately.
%If the number of sequences in the picture is k, the size of the picture is 28k $\times$ 392.
%Then one fifth of the text number of digital elements which has the size of 7$\times$7 pixels are added as noise to the random position in each image.
%The digital elements of the training set and test set are taken from the corresponding dataset in MNIST.
We generate 27,000 samples for training and 3,000 samples for testing for each dataset. % with $n=\{1, 2, 3, 4, 5\}$. Fig.\ref{fig:mlmnistl4} shows the sample of MS-MNIST[4].
\begin{figure}[htp]
	\begin{center}
		\subfloat[]{
			\includegraphics[width=0.8\linewidth]{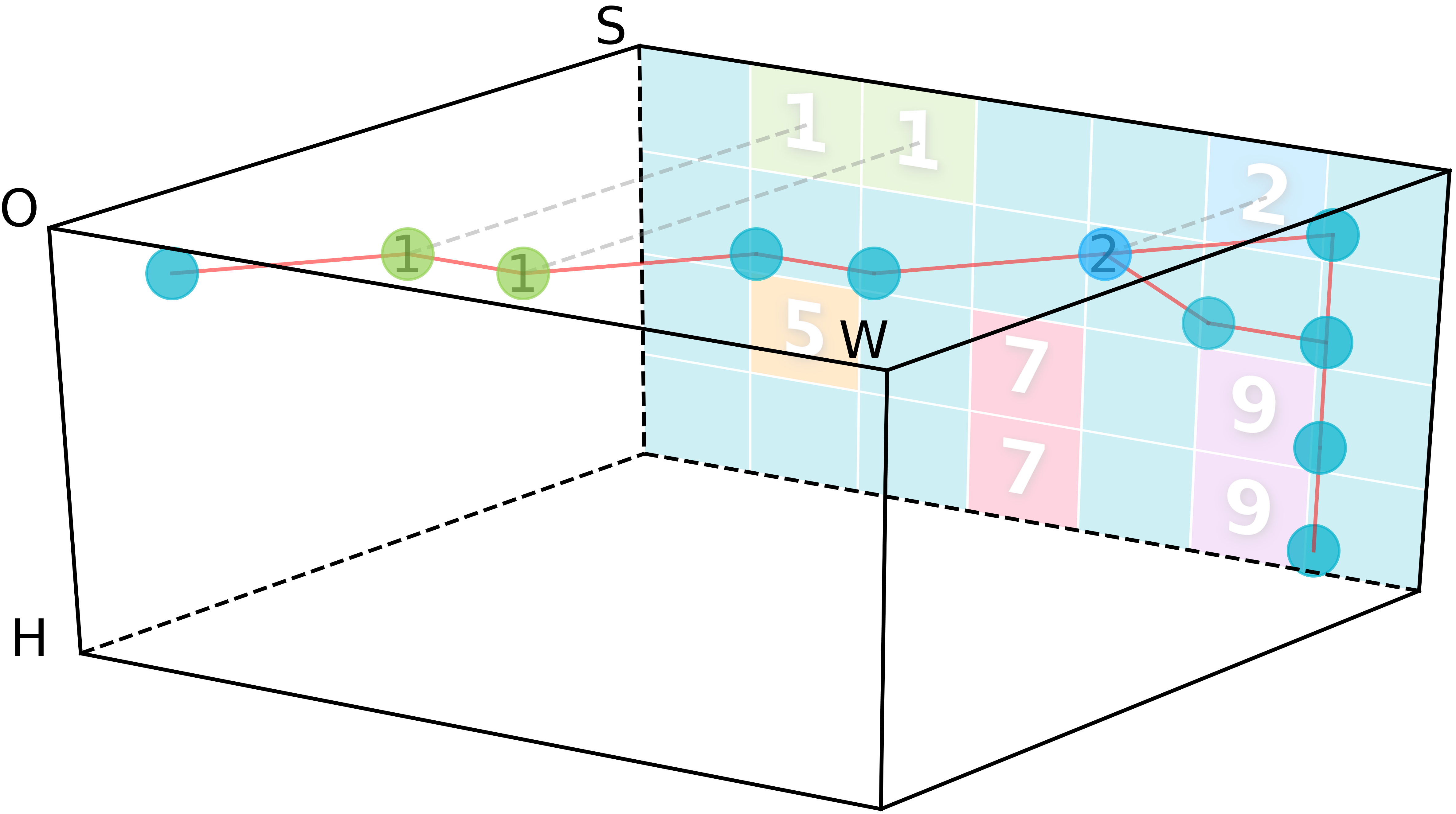}
		}\\
		\subfloat[]{
			\includegraphics[width=0.8\linewidth]{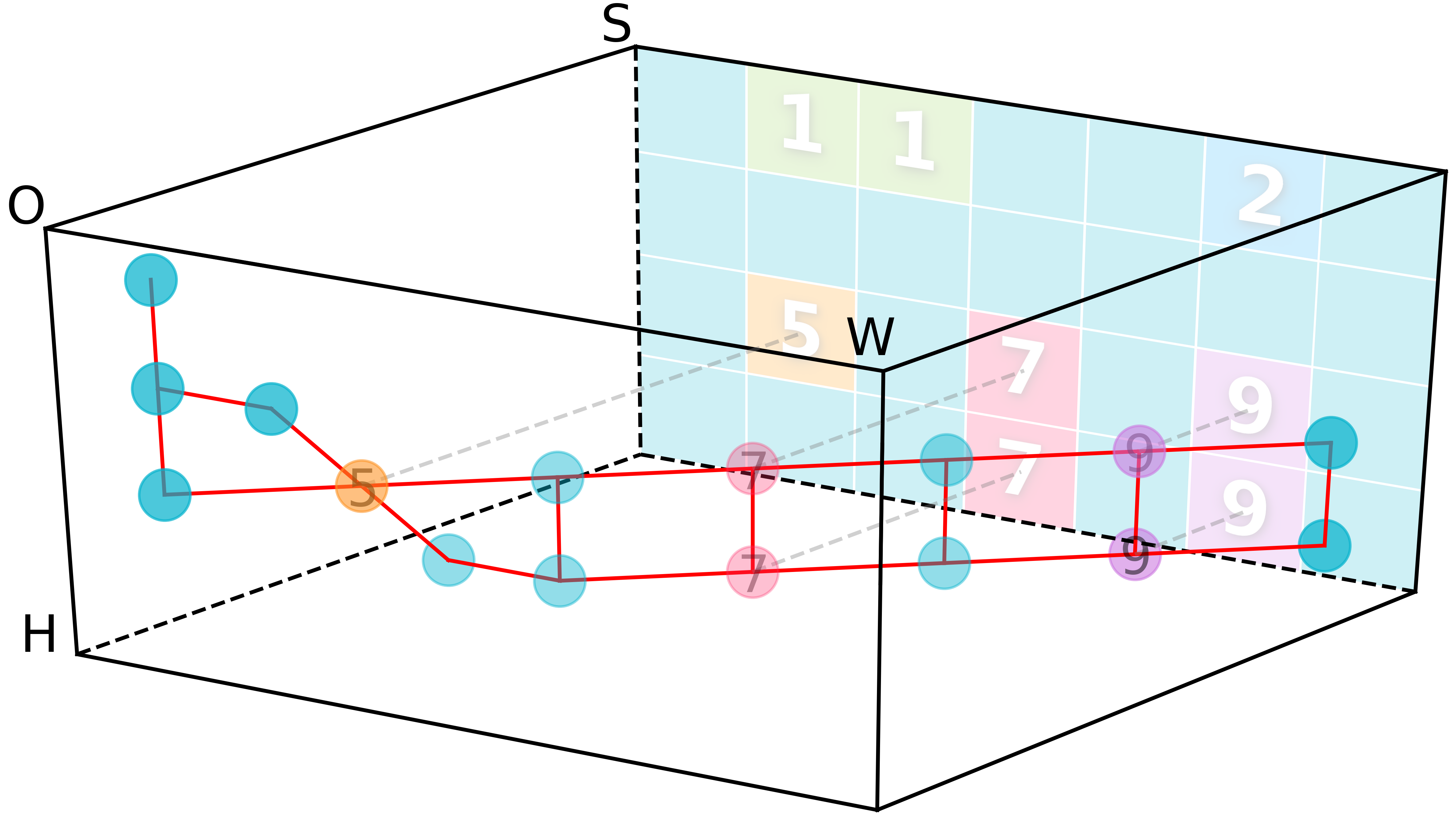}
		}
	\end{center}
	\caption{
		The illustration of decoding sequences (a) ``12'' and (b) ``579''.
		$O\rightarrow W$, $O\rightarrow H$ and $O\rightarrow S$ mean the width dimension, the height dimension and the length to be matched, respectively.
		Each filled circle means the corresponding class at $\alpha_{i,j}(s)$.
		%The point in the path represents the class of $\alpha_{i,j}(s)$ matching under the secondary state.
		%No digit points represent blank class.
	}
	\label{fig:pathinalpha}
\end{figure}

\subsubsection{Implementation details}
Our method is implemented under the framework of Caffe~\cite{Jia2014Caffe}.
We use 7 layers convolution structure (as in~\cite{shi2017end}) as the main network to extract image features, whose $\textit{\{kernel size, channels, stride, pad\}}$ are \{3,64,1,1\}, \{3,128,1,1\}, \{3,128,1,1\}, \{3,256,1,1\}, \{3,256,1,1\}, \{3,512,1,1\} and \{3,512,1,1\}, respectively.
Each convolution layer follows a {Relu}~\cite{Nair2010Rectified} activation layer.
There is a pool layer after the 1st, 2nd, 4th, 6th convolution layers, its kernel size is 2 $\times$ 2.
Then, three methods, 2 CTC/attention based baseline methods and our MSRA, are used to process the features, respectively.

MSRA and the CTC based method compute their respective constructed losses with the output of the softmax layer.
While the attention based method is implemented roughly (but not completely ) as in~\cite{bluche2017scan}.
Note that in our attention based method, it is necessary to re-calibrate the corresponding unordered sequence set from top to bottom and add a \emph{line break symbol} class between the sequences.

All networks are trained with the ADADELTA~\cite{zeiler2012adadelta} optimization method, and  $\lambda_1, \lambda_2$ are set to 0.9, 0.1.  
%$\sigma(x_1, x_2) = 0.9 x_1 + 0.1 x_2$.
Our experiments are performed on a workstation with an Intel Xeon(R) E5-2650 2.30GHz CPU, an NVIDIA Tesla P40 GPU and 128GB RAM.
\begin{figure*}[t]
	\begin{center}
		\subfloat[]{
			\includegraphics[width=0.25\linewidth]{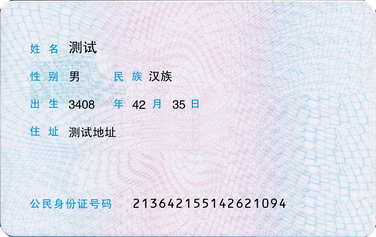}
		}
		\subfloat[]{
			\includegraphics[width=0.25\linewidth]{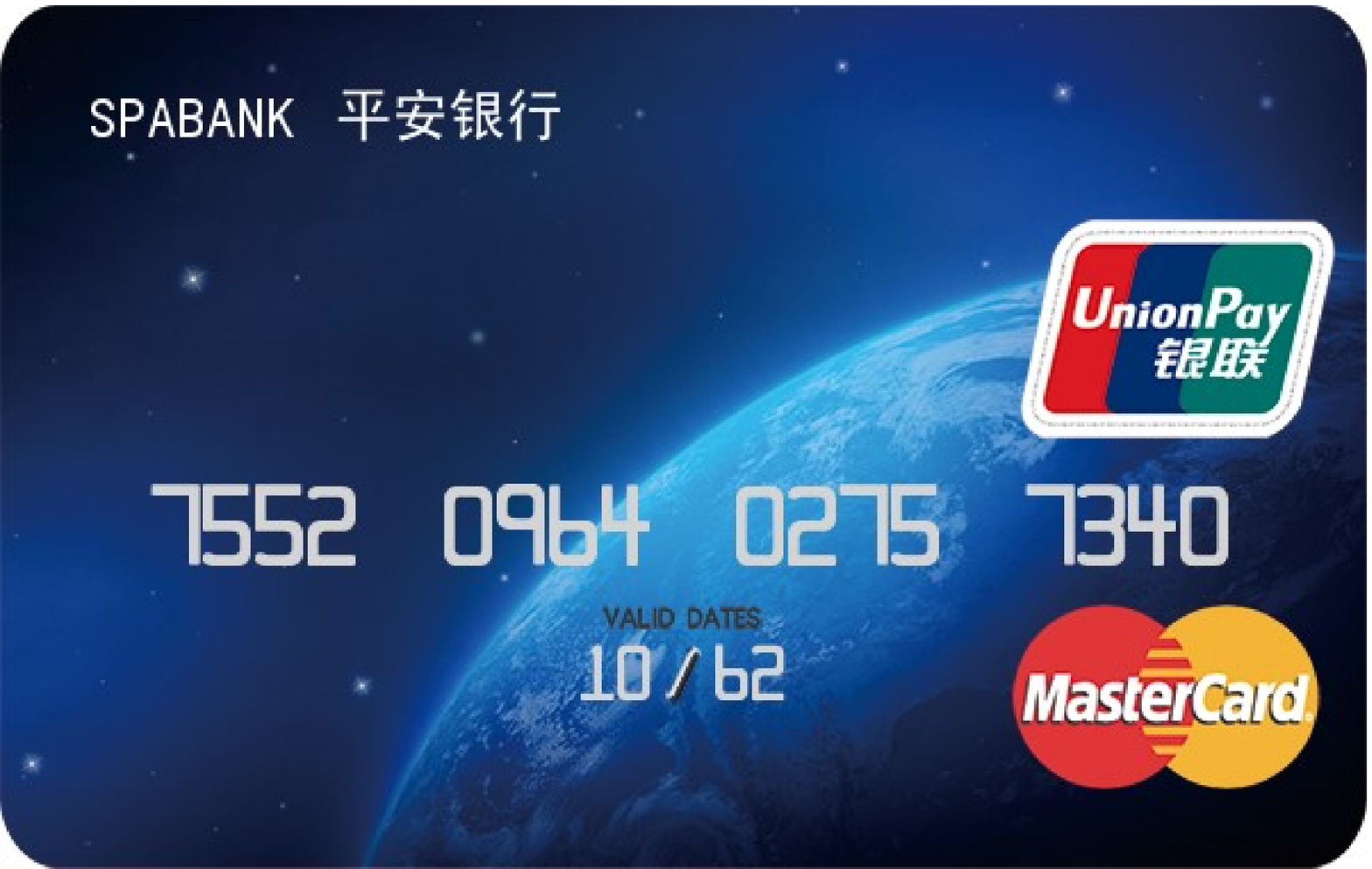}
		}
		\subfloat[]{
			\includegraphics[width=0.16\linewidth]{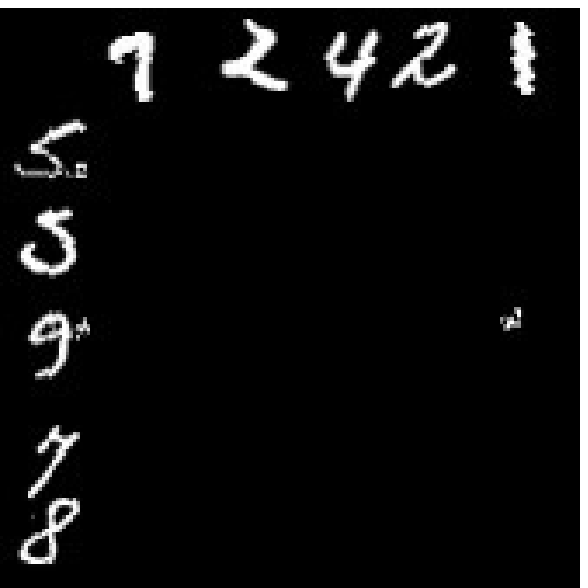}
		}
		\subfloat[]{
			\includegraphics[width=0.25\linewidth]{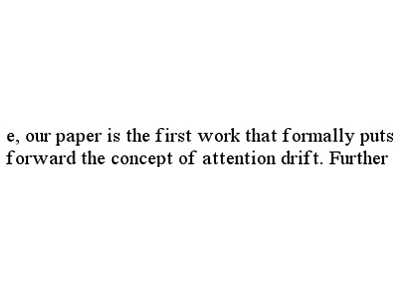}
			
		}	
	\end{center}
	\caption{
		Samples of four more challenging datasets: (a) IDN, (b) BCN, (c) HV-MNIST, and (d) SET.
	}
	\label{fig:real_dataset}
\end{figure*}

\subsubsection{Results}
Tab.~\ref{tab:mlmnist} shows the recognition results of the three methods mentioned above.
NED(\%) is normalized edit distance~\cite{marzal1993computation},
SA(\%) is sequence recognition accuracy,
and IA(\%) is image recognition accuracy.
As we can see from the table,
our MSRA method achieves better performance in all cases than the other two methods, and the CTC based method is unable to identify multiple text sequences in an image.
Though the attention based method shows acceptable performance when the number of sequences contained in images is small, but its performance degrades rapidly when the number of sequences contained in images increases to 5.

More importantly, MSRA can be trained with unordered sequence calibration, while the attention based method requires the text sequences in training images to be calibrated from top to bottom, which actually reveals the sequences' spatial layout information.
%, that is, multiple sequences in each picture are unrelated at the time of calibration.
%The ATTENTION method requires multiple sequences to be calibrated from top to bottom, which reveals part of the sequence location information.
When the attention based method uses disordered sequence calibration, the model can not converge. That is, it can not identify the situation.
As ordered calibration is a subset of disordered calibration, 
{{using ordered calibration in MSRA does not affect the results.}}
%Experiments also show that the convergence rate of POL method is faster than other methods.

%{\color{red}
We also visualize the matching paths for decoding text sequences in $\alpha$ space. Fig.~\ref{fig:pathinalpha} presents the visualization of prefix path matching probability $\alpha$ in an image containing two sequences ``12'' and ``579''.
Concretely,
Fig.~\ref{fig:pathinalpha}(a) and (b) separately show the matching process of the extended sequences ``-1-2-'' and ``-5-7-9-'' in $\alpha$ space. Here, `-' denotes the blank class.
In each sub-figure, the corresponding maximum probability matrix of conditional probability distribution is given on the back side of the cuboid.
The projection of each path along the $O \rightarrow S$ dimension is consistent with the result of the maximum probability class map.
Note that there are multiple paths for each target sequence.
For example, Fig.~\ref{fig:pathinalpha}(b), there are 2 and 6 possible paths before and after the `5' position,  %two paths before matching '5' and six paths after that.
which results in totally 12 possible matching paths.
\subsection{Performance on More Challenging Datasets}

In this part, we set up four more challenging datasets mainly based on real application scenarios, and train the MSRA model to recognize text sequences from images in these datasets, which are described in detail as follows:
\begin{itemize}
	\item \textbf{Identification card number dataset} (IDN).
	We create the \textbf{IDN} dataset by considering the standard font size and type used in the standard ID card.
	%add random simulation information to each corresponding position in the template strictly according to the font size and font type used by the standard ID card.
	Here, we just recognize the digital information in ID cards, including the ID card number, birthdate of the ID holder etc.
	%whose length are 18 and 8 respectively.
	\item \textbf{Bank card number dataset} (BCN).
	We generate the \textbf{BCN} according to the standard font size and type commonly used in 161 major banks.
	%Based on the bank card template, add random simulation information to each corresponding position in the template according to the font size and font type commonly used by major banks.
	We also just recognize the digital information including bank card number and valid period.
	%The bank card number is divided into two forms: 4-4-4-4(4 represents the number string length, "-" indicates the interval between strings) and 3-16.
	%And the length of the valid period is 4, of which 2 constitute the month and 2 consititue the date.
	\item \textbf{Horizontal and vertical MNIST dataset} (HV-MNIST).
	In this dataset, a horizontal sequence and a vertical sequence appear in each image.
	%They are located at the top and left side of the image respectively.
	Each sequence consists of 5 handwritten digits from MNIST {{and some noise just like in MS-MNIST datasets}}.
	%We need to adjust the selection gate parameters as $d1=0.5, d2=0.5$ to train the network.
	\item \textbf{Synthetic English Text dataset} (SET).
	We use 70 character-classes (26 capitals, 26 lowercase letters, 10 numbers and 8 other characters such as %\textbf{`,', `.', `:', `(', `)', `[', `]', ` '}
	`\textbf{,}', `\textbf{.}', `\textbf{:}', `\textbf{(}', `\textbf{)}', `\textbf{[}', `\textbf{]}', `\textbf{ }'
	) to generate multi-line English text based on some English documents downloaded from Web.
	The font is \textit{Times New Roman}.
	We randomly select a document and then randomly select a part of it as the label to generate image data.
\end{itemize}
Fig.~\ref{fig:real_dataset} shows some samples of the four datasets above, each of which contains 27,000 samples for training and 3,000 samples for testing.

For the four datasets, the network structure used in experiments is similar to that used for the MS-MNIST datasets.
The difference lies in the number and location of the pool layer and the reshape parameters of the input layer.
%The size of $\Gamma$ is controlled by changing the above variables.
Because we must ensure that $\mathbf{X}$ has the ability to hold multiple text sequences, in both horizontal and vertical directions.
For example, in the HV-MNIST dataset, we control the size of $\mathbf{X}$ to 14 $\times$ 14 for covering the extended length (11) of label strings, and some additional space to ensure that {{beginning}} step and the {{end}} step are blank.
%Because in this dataset, the maximum length of the horizontal and vertical text sequences is 5.
%It becomes to 11 after supplementing blank class.
%We also need to reserve some space to ensure that the initial step and the termination step are blank.
%Otherwise, each sequence will be forced to match the step's text information.
%When solving multiple line scene problems, we should consider not only information capacity but also receptive field.
%It is wrong to regard areas that do not observe effective information as part of the information capacity.
\begin{table}[!htb]
	\begin{center}
		\begin{tabular}{|c|c|c|c|}
			\hline
			\textbf{Datasets} & \textbf{NED} & \textbf{SA} & \textbf{IA} \\
			\hline
			\textbf{IDN} & 0.59 & 97.59 & 90.39 \\
			\hline
			\textbf{BCN} & 0.12 & 98.12 & 96.23 \\
			\hline
			\textbf{HV-MNIST} & 1.87 & 90.99 & 82.73 \\
			\hline
			\textbf{SET}& 1.48 & 68.57 & 47.90 \\
			\hline
		\end{tabular}
		\caption{Recognition results on datasets IDN, BCN, HV-MNIST and SET. NED, IA and SA are all evaluated in terms of percentage.}
		\label{tab:scene}
	\end{center}
\end{table}
\vspace{0.5cm}

Tab.~\ref{tab:scene} shows the recognition results of our method on the four datasets. We can see that our method still achieves promising performance.
MSRA performs satisfactorily on IDN and BCN though data in these datasets are based on real applications.
%These experiments have achieved good performance.
%Although their recognition targets are all two sequences, the results are different.
%Comparing results on IDN and BCN, we can see that MSRA is more suitable for reading text from images in BCN. This is because the complexity of identifying targets and the influence of background areas.
Results on HV-MNIST demonstrate that MSRA can handle complex MSR problems where an image has text sequences of different orientations.
%The simultaneous occurrence of sequences of different orientations can still be covered by paths in $\Gamma$, which theoretically shows that it can be recognized.
%The data in the table also confirm this point.
Our method has the lowest performance on SET, because  data in SET are more complex in terms of the number of classes and the length of sequences.
%Compared with MLmnistl2, the classes extends from 10 characters to 70 characters, and the length extends from 14 to 44.
The increase of class number means that for each step the method faces more matching options.
While the increase of sequence length means that a larger $\mathbf{X}$ is required to accommodate sequence information.
Though these datasets have more noise caused by complex background, different orientations, font size and type etc., MSRA still performs well.
%Its NED still has a good performance, and its SA is significantly lower than other scenarios mainly because of %its greatly increased sequence length.
%We also use different types of documents as sources to generate new test samples, which still maintain the existing recognition effect.

% !TeX spellcheck = <none>

%TODO:表 展示识别效果
%\section{Discussion}
%Multi-line text recognition problem is a scene in multi-sequence recognition, as shown in Figure x, multi-sequence recognition also contains more scenes.
%POL is a multi-sequence oriented method for the first time.
%We hope it can be used not only in multi-line text recognition scenarios, but also in more complex multi-sequence scenarios.
%多行文本识别问题时多序列识别中的一个场景，如图x所示，多序列识别还包含更多场景。
%POL是首次提出的面向多序列的方法，我们不仅希望它可以应用在多行文本识别场景，还可以应用在更加复杂的多序列场景中。

%In addition to text recognition, multi-sequence-oriented methods are promising in speech recognitions and other sequence-oriented problems.
%It saves the steps of sequence segmentation, reduces the workload of preprocessing and reduces the impact of segmentation anomalies.
%Our future work will continue to solve multiple sequence scene problems with multi-sequence-oriented methods.
%面向多序列的方法除文本识别外，在speech recognition等以序列为对象处理的问题上都大有可为。
%它会省去序列分割的步骤，减少预处理的工作量以及降低分割异常带来的影响。
%我们未来的工作会继续使用面向多序列的方法解决多序列场景问题。

\section{Conclusion}
In this paper, we address a novel and challenging problem, recognizing multiple text sequences from images by pure end-to-end learning. 
To this end, we propose an effective approach called MSRA. %, and evaluate our method several different challenging datasets.
Different from the existing methods, MSRA directly trains and identifies multiple text sequences in a pure end-to-end way without using any geometric annotations.
Experiments on several multiple text sequences datasets show that our MSRA method is effective and achieves better performance than two CTC and attention based baseline methods. % is better than the existing feasible scheme and its ability to solve practical problems.
In the future, on the one hand, we will conduct more comprehenisve performance evaluation on the MSRA method with pulic datasets; On the other hand, we will explore more advanced pure end-to-end techniques to solve the MSR problem.

{\small
	\bibliographystyle{ieee}
	\bibliography{egbib}
}

\end{document}